\documentclass[a4paper, 10pt, conference]{ieeeconf}      

\IEEEoverridecommandlockouts                              
\usepackage[utf8]{inputenc}
\usepackage[T1]{fontenc}
\usepackage{amsmath}
\usepackage{color}

\usepackage{graphics} 
\usepackage{epsfig} 
\usepackage{amssymb}  
\usepackage{multirow}

\title{\LARGE \bf
Socially Aware Crowd Navigation with Multimodal Pedestrian Trajectory Prediction for Autonomous Vehicles
}

\author{Kunming Li, Mao Shan, Karan Narula, Stewart Worrall, Eduardo Nebot* 
\thanks{*K. Li, M. Shan, K. Narula, S. Worrall, E. Nebot  are with the Australian Centre for Field Robotics (ACFR) at the University of Sydney (NSW, Australia). E-mails: {\tt\small \{k.li,m.shan,k.narula,s.worrall, e.nebot\}@acfr.usyd.edu.au}}
}

\begin{document}

\maketitle
\thispagestyle{empty}
\pagestyle{empty}

\begin{abstract}
Seamlessly operating an autonomous vehicles in a crowded pedestrian environment is a very challenging task. This is because human movement and interactions are very hard to predict in such environments. Recent work has demonstrated that reinforcement learning-based methods have the ability to learn to drive in crowds. However, these methods can have very poor performance due to inaccurate predictions of the pedestrians' future state as human motion prediction has a large variance. To overcome this problem, we propose a new method, SARL-SGAN-KCE, that combines a deep socially aware attentive value network with a human multimodal trajectory prediction model to help identify the optimal driving policy. We also introduce a novel technique to extend the discrete action space with minimal additional computational requirements. The kinematic constraints of the vehicle are also considered to ensure smooth and safe trajectories. We evaluate our method against the state of art methods for crowd navigation and provide an ablation study to show that our method is safer and closer to human behaviour.
\end{abstract}

\section{INTRODUCTION}
\begin{figure}
    \centering
	\includegraphics[width=6cm]{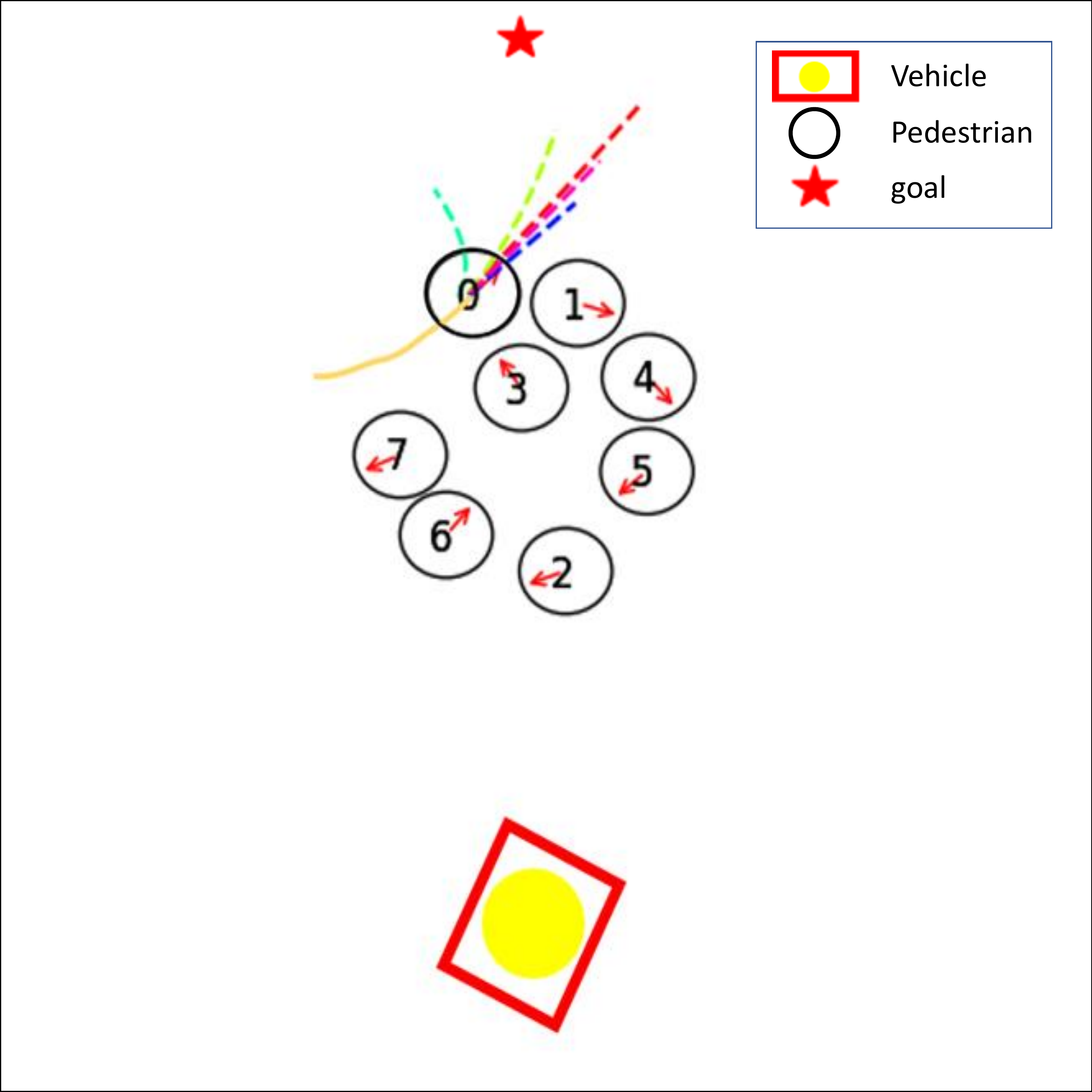}
	\caption{ Example of multimodal future trajectories for one pedestrian is shown from our experiment. The orange solid line is the observed trajectory, while the different color dash lines represent the pedestrian's future possible position. A arrow direction represents the walking direction of the pedestrian. Each pedestrian is represented by a circle with a number. The pedestrians walking in crowds may follow several plausible trajectories to avoid collisions.}
	\label{front_image}
\end{figure}

Autonomously driving in a crowded pedestrian environment is still a major challenge. There is a significant body of research examining this aspect of operating autonomous robots in a social environment with pedestrians. These tasks are difficult as they require vehicles or robots to have the ability to seamlessly and safely interact in close proximity to pedestrians. This problem becomes more complex when we consider uncertainty, the multimodal nature of human motion prediction as well as the implicit interaction among the crowds as shown in \textit{Fig. \ref{front_image}}. When a pedestrian is walking in crowds, he or she could develop several behavioral modes to avoid collision, and these modes have a large variance \cite{gupta2018social, kosaraju2019social}, increasing the difficulty of navigation. 

Prior work for vehicles or robots operating in crowded environments have treated pedestrians as dynamics obstacles with a simple kinematic model, or treat pedestrians as static obstacles and react using specific rules for avoiding collision \cite{fox1997dynamic, van2008reciprocal, phillips2011sipp}. These approaches have lead to unnatural and unsafe behaviour from the vehicles or robots. This is because traditional methods do not capture human motion and use only one-step look-ahead to react to pedestrians \cite{chen2019crowd, chen2017socially}.

There are a number of recent publications that have explored the aspects of pedestrian trajectory prediction and modelling of human-human interactions \cite{gupta2018social, scholler2020constant, alahi2016social}. These predictions have not, however, been used in the decision making process for planning in a crowded environment. Furthermore, some works have proposed to separate the autonomous task into prediction and planning, considering the problem as having two steps. The planner is used to select a safe path to avoid pedestrians at a given location. However, this can easily result in the \textit{freezing robot problem}, where vehicles find that all of the predicted paths are unsafe and freeze, or continually stop, in a crowded environment \cite{trautman2013robot}. To address the \textit{freezing robot problem}, existing works have proposed to provide vehicles with a set of cooperative rules when navigating in dense crowds. This however may lead to a high computational cost as demonstrated in \cite{chen2019crowd}.

Reinforcement learning has gained increasing popularity in automation tasks since it has been shown to produce paths that are close to human behavior. These paths can encode features related to the interactions and the cooperation between vehicles or robots \cite{chen2017socially, kuderer2012feature, chen2017decentralized, long2018towards}. These works have only had limited success, failing to fully comprehend the interaction between pedestrians, or to model cooperation between pedestrians and robots. Recently, some methods have looked to address this problem by extracting interaction features and inferring relative importance between the members of the crowd \cite{chen2019crowd, chen2019relational}. Although these methods have demonstrated promising results when operating in crowded environments, there are still some important limitations. Firstly, the methods are based on the unrealistic assumption that the vehicles can have access to either the ground truth of the future state of each pedestrian \cite{chen2019crowd} or predictions of a single potential trajectory \cite{chen2019relational}. This is unsafe for both vehicles and pedestrians as they do not consider the uncertainty of human motion prediction. These methods do not have motion constraints for controlling vehicles or robots, which is unrealistic and potentially dangerous in a real-world application. Besides, the action space has a limited number of actions, leading to movements of the vehicle that are potentially unnatural and unsafe.

In this work, we propose a novel algorithm to address the problem of operating an autonomous vehicle in highly crowded environments. This work will consider:
i) the prediction of future multimodal pedestrian trajectories incorporating interactions between pedestrians into a reinforcement learning network without retraining, and 
(ii) an extended action space with kinematic constraints, which are embedded to generate actions that enable the vehicle to drive seamlessly and safely. This is achieved with minimal additional computational requirements.
We also present a number of experiments to demonstrate the performance of the algorithm presented with comparisons to state-of-the-art methods.

\section{Related Work}

\subsection{Multimodal Pedestrian Trajectory Prediction}

Understanding human motion is critical for autonomous systems when they are to operate in crowded environments. This becomes increasingly complex when we consider the uncertainty and multimodality of pedestrian motion. Earlier work has largely leveraged a well-engineered model of forecasting human behavior. An example of prior work in this area is the human to human interaction model proposed by Helbing and Molnar \cite{socialforce}. This approach models pedestrian behavior and interactions with "social force" guiding them toward their destination while avoiding collisions with each other.

The social force concept has been revisited by \cite{choi2012unified, choi2013understanding, leal2014learning}. In this work, an Interaction Gaussian Process is proposed to model the trajectory of each pedestrian toward their goal as an individual Gaussian Process. The method is further extended to multi-goal by \cite{choi2014real} and applied in crowded environments \cite{trautman2015robot, trautman2017sparse}. In \cite{van2008reciprocal}, the authors introduced the Reciprocal Velocity Obstacle (RVO) concept, which is extended from the velocity obstacle concept for navigation among moving obstacles. Optimal Reciprocal Collision Avoidance (ORCA) proposed in \cite{van2011reciprocal} derives sufficient conditions for collision-free motion based on the velocity of obstacles. However, all these methods have the fundamental limitation that they rely on handcrafted energy potential and specific rules.

In recent years, learning-based methods have shown promising performance in predicting pedestrian motion. Recurrent Neural Networks are one of the popular learning-based methods recently used in forecasting pedestrian motion \cite{zhang2019sr}. Although these methods have obtained encouraging results, they have not considered the multimodal aspect usually present in predicting the future motion of pedestrians. The pedestrians in real world may follow a path from a set of likely trajectories, which can be represented as a multimodal distribution \cite{gupta2018social, kosaraju2019social, alahi2016social, robicquet2016learning}. 
This paper proposes a novel algorithm that incorporates multimodal pedestrian trajectory forecasting and vehicle kinematic constraints to assure seamless pedestrian-vehicle interactions, efficient operation and safety. 

\begin{figure*}[tbp] 
\centering
\includegraphics[width=\textwidth]{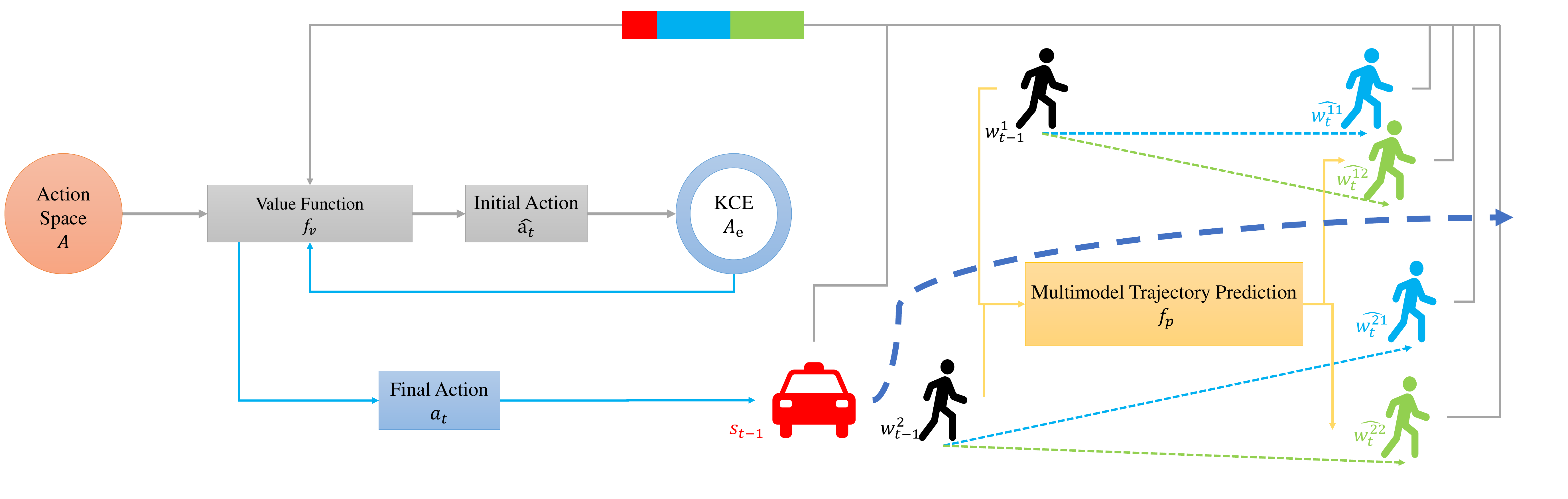}
\caption{Observed pedestrian trajectories are passed to the multimodel generator $f_{p}$, whilst the position of all agents, including the vehicle, are passed to a deep value function $f_{v}$ to find the initial action $\bar{a_{t}}$ with the highest value from action space $A$. The kinematic constraints and action space exploration (KCE) module outputs optimal action $a_{t}$ constrained by a kinematic model through value function $f_{v}$, from which action space is extended based on $\bar{a_{t}}$.}
\label{fig_archi}
\end{figure*}

\subsection{Operating in Crowded Pedestrian Environments}
Operating autonomous vehicles in highly crowded environments is challenging due to both the complexities of human motion, and interactions between pedestrians and vehicles. Prior work has approached this problem using dynamic models, such as \cite{van2008reciprocal, van2011reciprocal}. However, these methods rely on handcrafting a model to represent the control of the vehicle. Chen \textit{et al.} proposed CADRL, which aims to learn to drive when the pedestrian intention is unobserved using reinforcement learning \cite{chen2017decentralized}. They also proposed a method named SARL that captures the human-human interactions in the reinforcement learning framework and learns to drive in crowds by extracting the collective importance of neighbouring people\cite{chen2019relational, chen2019crowd}. However, their methods rely on a value function to find optimal actions by predicting a single trajectory for an individual pedestrian, or by using rule-based methods. This approach could cause the vehicles to collide with a pedestrian because of the uncertainty of human behaviour. LeTS-Drive \cite{cai2019lets} proposed an algorithm using an online belief-tree search. They utilize Convolutional Neural Networks to extract features from the environment as well as the intended path of vehicles, which may be unreliable \cite{chen2019relational}. Most methods presented use data from a simulator. The methods presented in this paper take advantage of learning data from real-world experiments, showing very promising performance.
\subsection{Reinforcement Learning}
Reinforcement learning has been widely applied in various fields including video games, control of robots, to name a few \cite{silver2017mastering, sutton1992reinforcement, gullapalli1994acquiring}.
Driving in crowded environments can be formulated as a decision-making problem under a reinforcement learning framework \cite{chen2017decentralized, chen2017socially}. Lei \textit{et al.} present a learning-based, mapless motion planner using deep reinforcement learning \cite{tai2017virtual}. A motion planner for crowded environments using reinforcement learning has been further revisited recently by \cite{chen2017decentralized, chen2019relational, chen2019crowd}, where Chen \textit{et al.} proposed a collision avoidance algorithm using deep reinforcement learning, which focuses on non-communicating multiagents. CADRL \cite{chen2017socially} specifies violations of social norms within a deep reinforcement framework. SARL \cite{chen2019crowd} and MP-RGL \cite{chen2019relational} consider social interaction between pedestrians by applying a graph neural network with pool mechanism in value function. However, those methods do not take the multimodality of pedestrian motion prediction into consideration, which could degrade the performance of the planner. Furthermore, although those methods select actions from a relatively small discrete action space to reduce computation, it restricts the agent (vehicle or robot) learning resulting in less natural trajectories. 

\section{Approach}
\subsection{Problem Formulation}
Driving in crowds can be regarded as a sequential decision making problem in reinforcement learning \cite{chen2017decentralized, chen2019relational, chen2017socially, chen2019crowd}. 
In this work we assume that each agent (a pedestrian or a vehicle) can observe position  $\textbf{P}=  [p_{x},p_{y}]$, velocity $\textbf{V} = [v_{x}, v_{y}] $  and radius (size) $ r $. The angular velocity of the agents at time $t$ is defined as $ \omega_{t} = [v_{s},\theta]$ . The goal position and preferred speed of vehicle are denoted as $\textbf{P}_{g}$ and $V_{pref}$. Let $a_{t}$ denote as actions as we assume that the velocity of vehicle $\textbf{V}_{t}$ will be achieved immediately once action $a_{t}$ is taken ($a_{t}=\textbf{V}_{t}$). We denote $s_{t} = [\textbf{P}_{t}, \textbf{V}_{t}, r, \textbf{P}_{g}]$ as the state of vehicle and $w_{t} = [w^{1}_{t},w^{2}_{t},w^{3}_{t}....,w^{n}_{t}]$ as the state of $n$ pedestrians. $\textbf{P}_{t}$ and $\textbf{V}_{t}$ in $s_{t}$ is defined as the position and velocity of the vehicle at time $t$. For each pedestrian $i$, his or her state is defined as $ w^{i}_{t}= [\textbf{P}_{t}, \textbf{V}_{t}]$, where $\textbf{P}_{t}$ is defined as the pedestrian's position at time $t$ and $\textbf{V}_{t}$ is defined as pedestrian's velocity time $t$. The joint state for this operation is defined as $s^{j}_{t}= [s_{t}, w_{t}]$ at time $t$. The optimal policy $\pi^{*}$ is to identify the action with maximum value, which is denoted as $\pi^{*}: s^{j}_{t} \mapsto a_{t} $ and detailed in \textit{Sec. \ref{whole_algo}}.


\subsection{Overview}
Our approach consists of applying an LSTM-based multi-model trajectory predictor. A deep value function captures the features of human-human and human-vehicle interactions. It also includes a module to incorporate the kinematic constraints in a deep reinforcement learning framework, as shown in \textit{Fig. \ref{front_image}}. In our work, Social GAN proposed by \cite{gupta2018social} is employed to estimate the future position of pedestrians. This algorithm generates a multi-model estimate of the future pedestrian trajectories by combining tools from a recurrent sequence-to-sequence model, generative adversarial networks and pooling module. The deep value function used for estimating actions of driving in crowded environments is proposed by \cite{chen2019crowd}. The actions of a vehicle are evaluated, considering the kinematic constraints through a reward function as well as a filter for removing unwanted actions. The action space is extended based on the initial optimal action, and the final optimal action is taken from an extended action space.

\subsection{Multimodal Trajectory Prediction} \label{whole_algo}

\subsubsection{Multimodal Trajectory Prediction} In this work, we employ Social GAN \cite{gupta2018social} to predict socially plausible features. It also generates different predictions by training adversarially against a recurrent discriminator. The vehicle firstly observes recent trajectories of all pedestrians in the scene and predicts multiple future likely trajectories as follows:

\begin{align}
    \hat{w_{t}} = f_{p}(w_{t-1}) \\
    \hat{w_{t}}= [\hat{w}_{t}^{11}, \hat{w}_{t}^{12},...,\hat{w}_{t}^{1m},...,\hat{w}_{t}^{nm}]
    \label{eq:predictor}
\end{align}
where $f_{p}$ is the predictor and $w_{t}$ are the states of all pedestrian within a scene. $\hat{w_{t}}$ represents multiple possible future trajectories for all pedestrians at time $t$ where each pedestrian has $m$ possible future trajectories.

\subsubsection{Deep Socially Attentive Value Function Network} A deep socially attentive value function network of SARL is proposed by \cite{chen2019crowd}. This function calculates the relative importance and encodes the collective impact of neighbouring pedestrians or robots for estimating socially compliant navigation actions.  In this work, we combine deep socially attentive value function of SARL with multimodal future trajectories produced by \cite{gupta2018social} to further improve the performance of the navigation algorithm. It is important to remark that Social GAN \cite{gupta2018social} generates multiple future likely trajectories for each pedestrian, reducing the search space for a vehicle when searching for potential collisions. Inspired by \cite{chen2019relational, chen2019crowd}, the value function of our work is formed as follows:
\begin{align}
V(\hat{a_{t}},\hat{w}_{t},s^{j}_{t-1}) = \operatorname*{argmax}_{\hat{a_{t}} \in \textbf{A}} (V_{m}(\hat{w}_{t},\hat{a_{t}},s_{t-1})) 
 \label{eq:value_function}
\end{align}

\begin{align}
V_{m}=R(\hat{w_{t}},\hat{a_{t}},s^{j}_{t-1})
+\gamma^{v_{pref}}V_{s}(\hat{w_{t}},s_{t-1},\hat{a_{t}})
\end{align}

The vehicle chooses an initial optimal policy by estimating a value of action $\bar{a_{t}}$ from the initial action space ($\textbf{A} = A_{i}$) using the value function $V_{m}$. $V_{s}$ is value function proposed by SARL \cite{chen2019crowd}. In this work, we estimate $\hat{a_{t}}$ through the multimodal trajectories prediction $\hat{w}_{t}$ and a potential vehicle future state calculated by $\hat{a_{t}}$ and $s_{t-1}$. The final optimal action $a_{t}$ will be selected from the expanded action space $A_{e}$. This space is based on the initial optimal action $\hat{a_{t}}$, as detailed in  \textit{Sec \ref{act_space_exp}}. $\gamma$ is defined as the discount factor.
Different from the reward function presented in \cite{chen2019crowd}, we propose a novel reward function with the vehicle kinematic constraints detailed in \textit{Sec \ref{kin_cons}} . 

\subsection{Incorporating Kinematic Constraints} \label{kin_cons}
We include more realistic kinematic constraints, as any robot will have inherent restrictions to their motion. \cite{augugliaro2012generation, chen2019crowd, socialforce} stated the potential complication of adding these capabilities in terms of additional computational complexity. In this work, we incorporate kinematics constraints into the reinforcement learning framework by modifying the reward function and action space. The potential actions are selected from a range of accelerations, velocities and turning angles, which narrow down the action space by calculating absolute difference in angular velocity between $\omega_{t-1}$ and $\omega_{t}$ as follows:

\begin{align}
    | \omega_{t-1} - \omega_{t} | \leq \Delta \omega_{max} \\
    \Delta \omega_{max} = [v_{smax},\theta_{max}]
\end{align}

A reward function $R$ is designed to reward the vehicle for reaching its goal with a smaller value of acceleration while penalizing collision,

\begin{figure*} 
    \centering
	\includegraphics[width=\textwidth]{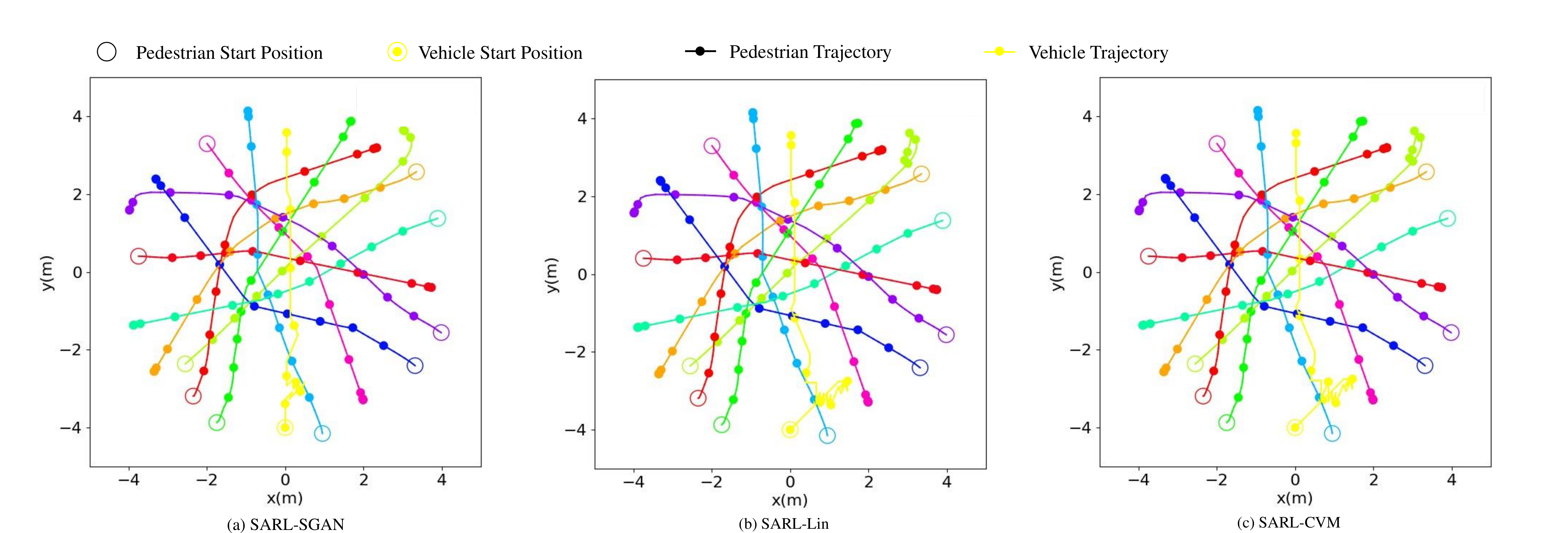}
	\caption{Comparison of tested methods for a 1-timestep action prediction in a complex environment. 10 pedestrians are shown in the simulator starting in a circle before crossing to the other side. It is clearly observed that SARL-Lin and SARL-CVM have frequent stops and show conservative vehicle behaviour when close to pedestrians.}
	\label{ex1_pdf}
\end{figure*}

\begin{equation}
R(\textbf{s}^{j}_{t},\textbf{a}_{t})=\left\{
\begin{aligned}
-0.25 + f_{\Delta}(a_{t},a_{t-1}) && {d_{min}<0}  \\
-0.1 - d_{min}/2 + f_{\Delta}(a_{t},a_{t-1}) && {d_{min}<0.2} \\
1 + f_{\Delta}(a_{t},a_{t-1}) && {\textbf{P}=\textbf{P}_{g}} \\
0 && {o.w}
\end{aligned}
\right.
\end{equation}

\begin{align}
     f_{\Delta}(a_{t},a_{t-1}) = threshold - |a_{t}-a_{t-1}|
\end{align}

where $d_{min}$ is the minimum separation distance between the vehicle and the humans during the period $[t-1,t]$

\subsection{Action Space Exploration} \label{act_space_exp}
The action space is narrowed down because the kinematic constraints are used to limit the subsequent searching for the optimal action. However, as the actions in action space are discrete, the difference between $a_{t}$ and $a_{t-1}$ maybe still large. We further refine the optimal action $\bar{a_{t}}$ by extending action space as below:
 
\begin{align}
A_{e} = \left\{y : \left(\exists m \in \mathbb{N}\right)
\left[
m \leq n \wedge y = \bar{a}_{t} \pm m\epsilon
\right]\right\},
\end{align}
where $\epsilon$ and $n \in \mathbb{N}$ are characteristics used to define the range of the action space. After generating the new action space $A_{e}$, the vehicle will go through \textit{ Eq. \eqref{eq:value_function}}, where $\textbf{A} = A_{e}$ and select the final action from $A_{e}$.

\section{Experiments}
\subsection{Experimental Setup}
We use the CrowdNav simulation environment provided by \cite{chen2019crowd} for the vehicle navigation in crowds.  The movement and interaction between pedestrians and vehicles are based on ORCA \cite{socialforce} and the parameters are sampled from Gaussian distributions. In this experiment, the models we tested are the circle crossing scenarios with 5 and 10 pedestrians. The pedestrians are randomly positioned around a circle of 4$m$ radius. Perturbations are added to the position of pedestrians, goal position and vehicle start position. All the models are evaluated with 100 Monte Carlo test cases, apart from the experiment to test methods using the 5 pedestrian scenarios in \textit{Experiment 1}. In the experiment \textit{Experiment 1}, we test the methods with 500 Monte Carlo test cases to compare with other existing methods. 

The Social GAN model \cite{gupta2018social}  used in the experiments is trained with a real-world dataset, ETH \cite{sadeghiankosaraju2018trajnet}. We use the pre-trained model of SARL from \cite{chen2019crowd}.

\subsection{Evaluation Metrics and Baselines}
\subsubsection{Baseline Comparisons}
As stated in \cite{chen2019relational}, SARL proposed by \cite{chen2019crowd} is based on the unrealistic assumption that the next ground state of all participant is available. We replace this assumption with two well-known predictors and compare their performance. One of them is the linear model motion prediction (Lin).  Lin has been widely used in navigation tasks and pedestrian trajectories prediction tasks \cite{chen2019relational}. The other model for comparison is the Constant Velocity Model (CVM) \cite{scholler2020constant}. In many cases, CVM has been shown to provide consistent performance in more simple motion prediction scenarios.

We compare our model against the following baseline:
\begin{itemize}

    \item ORCA: A dynamic model for operating robots in the crowds proposed by \cite{van2008reciprocal} .
    
    \item CADRL: A approach for driving in crowded environments socially based on reinforcement learning proposed by \cite{chen2017decentralized}.
    
    \item  LSTM-RL: a reinforcement learning based approach using LSTM to learn to make driving decisions proposed by \cite{everett2018motion}. 
    
    \item SARL-CVM: SARL with a constant velocity model for pedestrians future state predictions, which observes the last two points of trajectory and predict the next 8 points trajectory in the future.
    
    \item SARL-Lin: SARL with a linear model for pedestrians future state predictions, which observes 8 points of trajectory and predicts the next 8 trajectory points in the future.

    \item SARL-SGAN-v$n$: SARL with $n$ possible future trajectories generated by Social GAN \cite{gupta2018social}. Social GAN observes 8 points of the trajectory and predict 20 possible locations in the future. We define SARL-SGAN-v20 as SARL-SGAN by default. 
\end{itemize}

\begin{table}[h]
\centering
\vspace{3mm}
\caption{Quantitative results of tested methods for predicting 1-timestep action in the scenario with 5 pedestrians. We compare the results using the metrics of success rate and driving time from start position to goal position. The result of the success rate of ORCA \cite{van2008reciprocal}, CADRL \cite{chen2017decentralized} and LSTM-RL \cite{everett2018motion} are from \cite{chen2019crowd} while we ignore the results of time as they have different ``Time'' metrics from ours}
\label{ex1_ped5}
\begin{tabular}{lll}
\hline
\textbf{Method} & Success & Time  \\ \hline
ORCA \cite{van2008reciprocal} & 43\%  & Nil \\
CADRL \cite{chen2017decentralized} & 78\% & Nil \\
LSTM-RL \cite{everett2018motion} & 95\%  & Nil \\
SARL-Lin        & 97.2\%  & 12.58 \\
SARL-CVM        & 98.6\%    & 11.6 \\
SARL-SGAN (ours) & 99.2\%  & 11.7 \\ \hline
\end{tabular}
\end{table}

\subsubsection{Metrics}
We evaluate the models using the metrics as follows:

\begin{itemize}
    \item Success: The rate of vehicle reaching its goal without colliding with pedestrians within certain time.

    \item Time: The time duration a vehicle takes to move from starting point to the goal. 

     \item Maximum Acceleration: The maximum acceleration of the vehicle between start point and goal.
\end{itemize}

\begin{figure*}
    \centering
	\includegraphics[width=\textwidth]{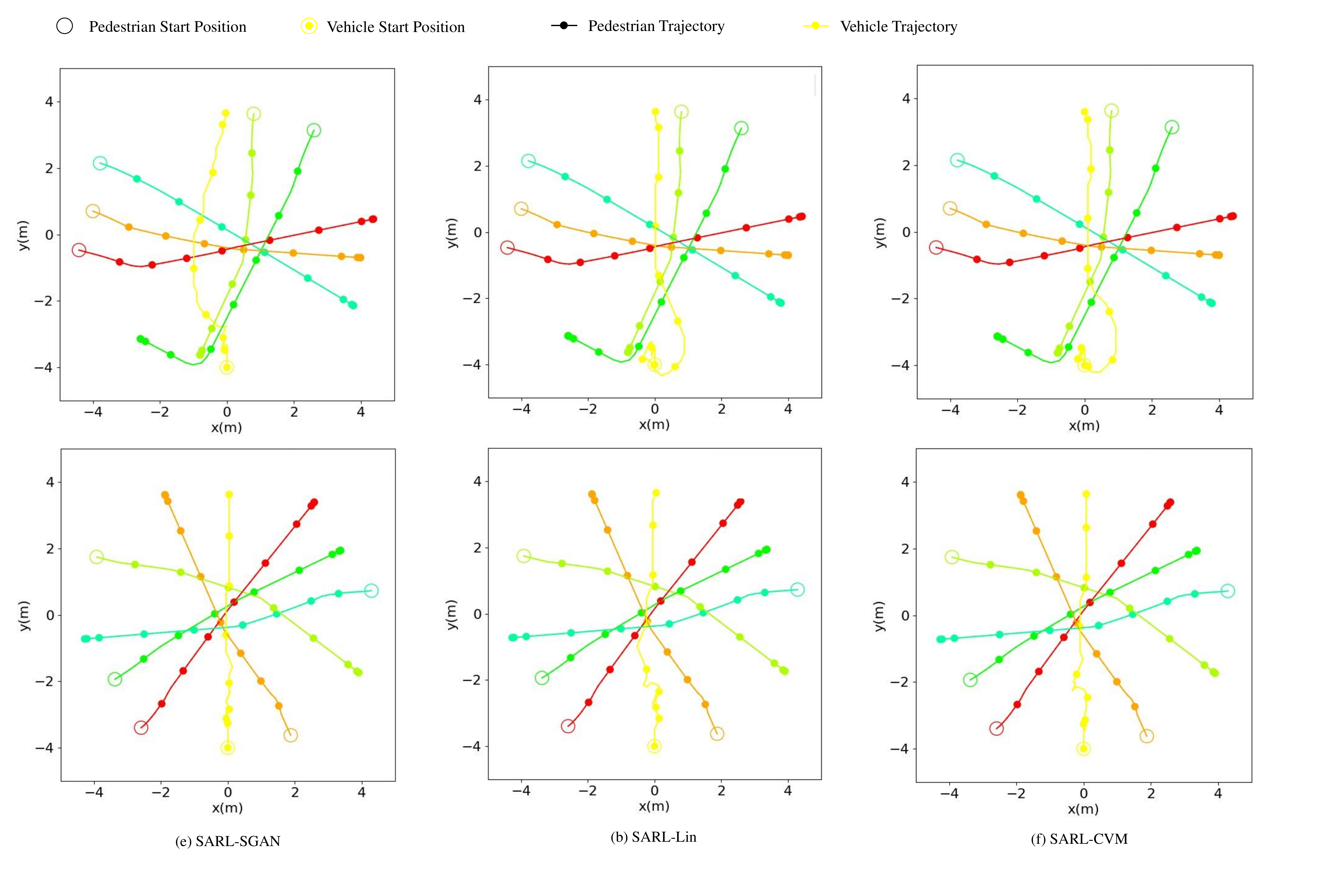}
	\caption{Comparison of tested methods for an 8-timestep action prediction in the simulator with 5 pedestrian. SARL-SGAN successfully identifies a shortcut while SARL-Lin and SARL CVM cannot predict this and both react conservatively.}
	\label{ex2_pdf}
\end{figure*}

\subsection{Methodology}
For the evaluation of our proposed method SARL-SGAN-KCE, we compare with the state of art methods and perform an ablation study through experiments as follows:

\textbf{Experiment 1.} We compare SARL-Lin, SARL-CVM, OCRA\cite{van2008reciprocal} and CADRL \cite{chen2017decentralized} with our proposed method without KCE (SARL-SGAN) to predicting an action for each timestep. We test the models in a simulator with 5 and 10 pedestrians. SARL-Lin, SARL-CVM and SARL-SGAN observe the last 8 timesteps of each trajectory (1.75 seconds).

\textbf{Experiment 2.} We further verify our methods by comparing them by predicting consecutive actions for multiple timesteps. We observe the previous 8 timesteps from each trajectory and select the actions for the next 8 timesteps.

\textbf{Experiment 3.} We evaluate the effectiveness of the approach with the kinematic constraints and action space exploration module (SARL-SGAN-KCE) using an ablation study.

\begin{figure*}
    \centering
	\includegraphics[width=14cm]{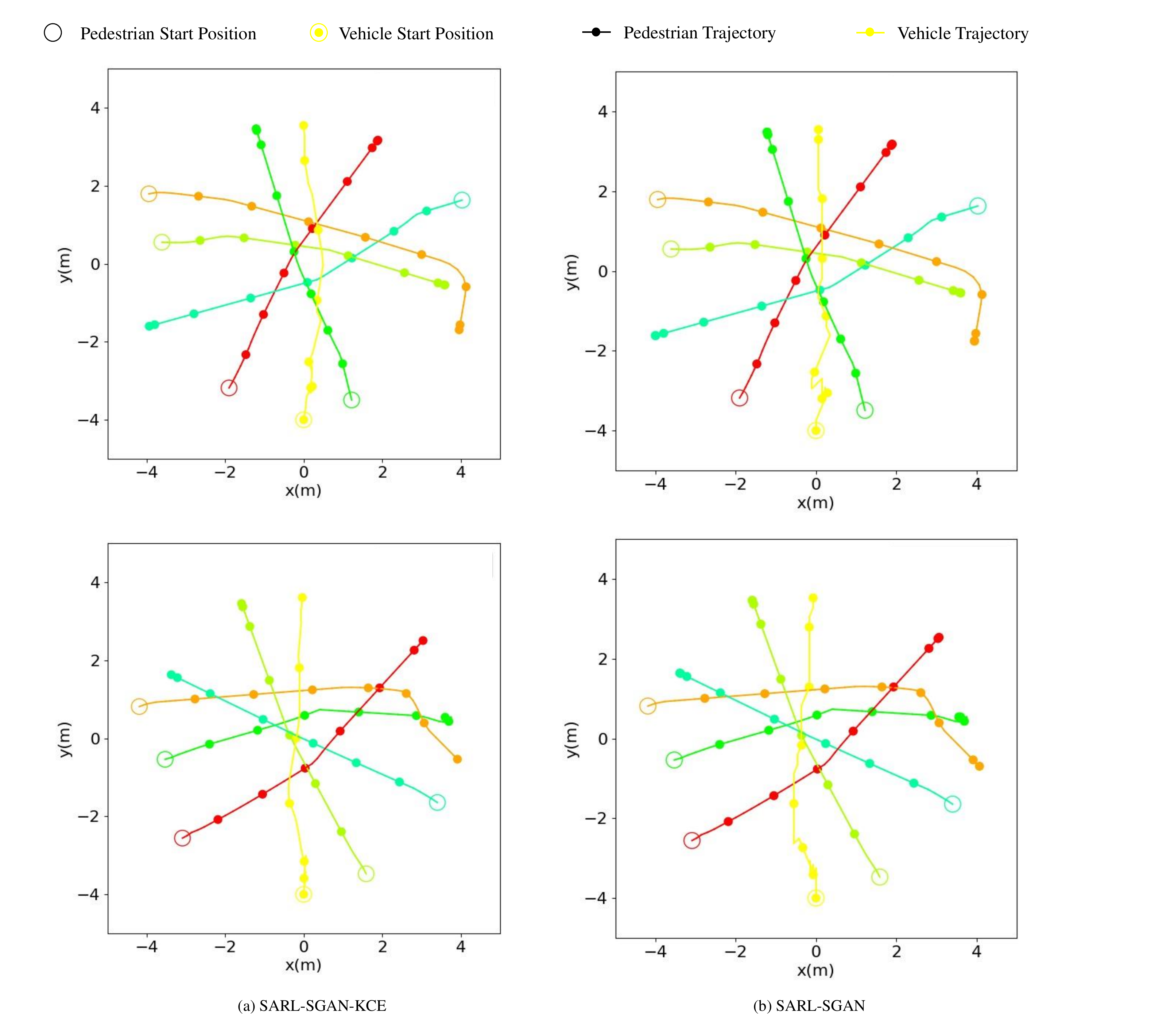}
	\caption{The SARL-SGAN-KCE has obviously smoother trajectory than the SARL-SGAN. The SARL-SGAN tends to stop in many cases. In contrast, the trajectory of SARL-SGAN-KCE looks natural and closer to human behaviour.}
	\label{ex3_pdf}
\end{figure*}

\section{Results}
\subsection{Quantitative Result} \label{ex1}

\begin{table}[]
\centering
\vspace{3mm}
\caption{Quantitative result of tested methods for predicting 1-timestep action in the scenario with 10 pedestrians.}
\label{ex1_ped10}
\begin{tabular}{lll}
\hline
\textbf{Method}            & Success               & Time                   \\ \hline
SARL-Lin                  & 97\%                  & 14.4                   \\
SARL-CVM                                       & 99\%                  & 13.55                  \\
SARL-SGAN-v5 (ours)                      & 99\%                  & 13.66                     \\ 
SARL-SGAN-v20 (ours)                      & 99\%                  & 13.61                     \\ \hline
\end{tabular}
\end{table}

\textbf{Experiment 1.} We first compare SARL-SGAN with other methods in scenarios with 5 and 10 pedestrians respectively as shown in \textit{Table \ref{ex1_ped5}} and \textit{Table \ref{ex1_ped10}}. From \textit{Table \ref{ex1_ped5}}, it can be shown that SARL-SGAN outperforms all other methods with a higher success rate, but a similar travelling time. As stated in\cite{chen2019crowd}, ORCA performs poorly in the scenarios where pedestrians are not aware of the state of vehicles, as it violates the reciprocal assumption. 
We further evaluate the models in scenarios with 10 pedestrians. Our proposed approach still outperforms the other methods. CADRL only takes a single pair of interactions into consideration, which is believed to be a major cause of a lower success rate. The SARL-SGAN has only a slight improvement over SARL-Lin, SARL-CVM, and LSTM-RL in this experiment because all methods are only considering 1-timestep prediction. All predictors perform nearly equally in the short-term prediction of future positions.

\textbf{Experiment 2.} In this experiment, we evaluate the robustness of the navigation algorithms. We compare the models by evaluating predictions of multiple future actions. Each model predicts 8 timesteps of future consecutive actions instead of 1-timestep action from the previous experiment.
As \textit{Table \ref{table_ex2}} shows, SARL-SGAN with multimodal trajectory prediction outperforms other methods with a higher success rate and an acceptable travelling time. The limitation for the SARL-Lin and SARL-CVM \cite{scholler2020constant} is that they do not consider the uncertainty of pedestrian motion. A single hypothesis for future trajectory is considered inadequate as a pedestrian may follow several plausible trajectories \cite{gupta2018social, kosaraju2019social}. Through the experiment, we can see that the success rate increases with an increased number of possible prediction trajectories.

\begin{table}[h]
\centering
\vspace{3mm}
\caption{Quantitative result of tested methods for predicting 8-timestep action in the scenario with 5 pedestrians.}
\label{table_ex2}
\begin{tabular}{lll}
\hline
\textbf{Method} & Success & Time  \\ \hline
SARL-Lin        & 96\%    & 13.35 \\
SARL-CVM        & 92\%    & 13.67 \\
SARL-SGAN-v5 (ours)    & 95\%    & 13.8  \\
SARL-SGAN-v10 (ours)   & 98\%    & 13.8  \\
SARL-SGAN-v20 (ours)   & 100\%   & 14.5  \\ \hline
\end{tabular}
\end{table}

\textbf{Experiment 3.}  We evaluate the SARL-SGAN-KCE by an ablation study in this experiment. Although previous experiments have shown that our proposed SARL-SGAN has obtained promising performance driving in crowded environments, it is not suitable for autonomous driving due to allowing of a large maximum acceleration in SARL-SGAN. In the real world, it is extremely dangerous and physically impossible for vehicles to decelerate from their maximum velocity to zero in a short time. In the experiment, we set the maximum acceleration to 6.4 $m/s^{-2}$ and the maximum turning angle to 120$^\circ$ within a timestep. With kinematic constraints incorporated and using action space exploration, we achieve a similar result of success rate but see a significant decrease in the maximum acceleration by 41\%, and the travelling time by 7\%.
\begin{table}[h]
\label{ex3_kc}
\centering
\vspace{3mm}
\caption{Quantitative result of SARL-SGAN with and without KCE in the scenario with 5 pedestrians. We use the result of SARL-SGAN form SARL-SGAN-V20 of Table \ref{ex1_ped5}.}
\begin{tabular}{llll}
\hline
\textbf{Method}    & Success Rate & Time  & Max Acc \\ \hline
SARL-SGAN-KCE & 98\%         & 10.94 & 1.16                 \\
SARL-SGAN         & 99\%         & 11.85 & 1.96                 \\ \hline
\end{tabular}
\
\end{table}
\subsection{Qualitative Results}
\textbf{Experiment 1:} We further investigate how SARL-SGAN can handle the more complex environment. \textit{Fig. \ref{ex1_ped10}} demonstrates SARL with different predictors in complex scenarios, where ten pedestrians are randomly positioned on a circle, and their goal position is the opposite side of the same circle. We can find that SARL-SGAN tends to find a path that directly leads to the goal. SARL with linear predictor and CVM tend to go in a different direction from the goal as they make inaccurate predictions of pedestrians' future state.  

\textbf{Experiment 2:} As for multiple timestep actions planing, compared with SARL-Lin and SARL-CVM, SARL-SGAN demonstrates its superior performance in finding a shorter way to the goal. At the beginning, it predicts multiple possible future positions of each pedestrian enabling this method to find a better solution. SARL-Lin and SARL-CVM tend to go to the direction opposite to the goal due to the less accurate single trajectory pedestrian prediction.

\textbf{Experiment 3:} To further investigate the kinematics constraints and action space exploration, we compare SARL-SGAN with and without KCE as \textit{Fig. \ref{ex3_pdf}} shows.
Comparing with the SARL-SGAN, the trajectory of SARL-SGAN-KCE looks more natural. The motion is constrained to limit the acceleration and turning rate, more closely resembling an Ackermann steered vehicle model.


\section{CONCLUSION}
In this work, we presented a novel deep reinforcement learning-based crowd navigation algorithm for autonomous driving system with multimodal pedestrian trajectory prediction. In particular, our proposed method allows autonomous vehicles to learn to drive safely and efficiently in a crowded environment, taking into account uncertainty in the pedestrians' motion. Current reinforcement learning-based methods are limited by a small number of discrete actions space due to the expensive computational requirements. This paper introduced a module to augment the action space at a minimal additional computational cost. We also introduced kinematic constraints into the algorithm to achieve more natural trajectories and seamless interaction with pedestrians. Our proposed method delivers promising performance and outperforms other state-of-the-art methods under different metrics with more natural driving behaviour.




\section*{ACKNOWLEDGMENT}
This  work  has  been  funded  by  the  Australian  Centre  for
Field Robotics(ACFR), University of Michigan / Ford Motors Company Contract ``Next generation Vehicles”,  Transport for New South Wales (TfNSW) and iMOVE CRC and supported by the  Cooperative  Research  Centres  program,  an  Australian
Government initiative.


\bibliographystyle{IEEEtran}  
\bibliography{egbib}

\begin{thebibliography}{10}
\providecommand{\url}[1]{#1}
\csname url@samestyle\endcsname
\providecommand{\newblock}{\relax}
\providecommand{\bibinfo}[2]{#2}
\providecommand{\BIBentrySTDinterwordspacing}{\spaceskip=0pt\relax}
\providecommand{\BIBentryALTinterwordstretchfactor}{4}
\providecommand{\BIBentryALTinterwordspacing}{\spaceskip=\fontdimen2\font plus
\BIBentryALTinterwordstretchfactor\fontdimen3\font minus
  \fontdimen4\font\relax}
\providecommand{\BIBforeignlanguage}[2]{{%
\expandafter\ifx\csname l@#1\endcsname\relax
\typeout{** WARNING: IEEEtran.bst: No hyphenation pattern has been}%
\typeout{** loaded for the language `#1'. Using the pattern for}%
\typeout{** the default language instead.}%
\else
\language=\csname l@#1\endcsname
\fi
#2}}
\providecommand{\BIBdecl}{\relax}
\BIBdecl

\bibitem{gupta2018social}
A.~Gupta, J.~Johnson, L.~Fei-Fei, S.~Savarese, and A.~Alahi, ``Social gan:
  Socially acceptable trajectories with generative adversarial networks,'' in
  \emph{Proceedings of the IEEE Conference on Computer Vision and Pattern
  Recognition}, 2018, pp. 2255--2264.

\bibitem{kosaraju2019social}
V.~Kosaraju, A.~Sadeghian, R.~Mart{\'\i}n-Mart{\'\i}n, I.~Reid, H.~Rezatofighi,
  and S.~Savarese, ``Social-bigat: Multimodal trajectory forecasting using
  bicycle-gan and graph attention networks,'' in \emph{Advances in Neural
  Information Processing Systems}, 2019, pp. 137--146.

\bibitem{fox1997dynamic}
D.~Fox, W.~Burgard, and S.~Thrun, ``The dynamic window approach to collision
  avoidance,'' \emph{IEEE Robotics \& Automation Magazine}, vol.~4, no.~1, pp.
  23--33, 1997.

\bibitem{van2008reciprocal}
J.~Van~den Berg, M.~Lin, and D.~Manocha, ``Reciprocal velocity obstacles for
  real-time multi-agent navigation,'' in \emph{2008 IEEE International
  Conference on Robotics and Automation}.\hskip 1em plus 0.5em minus
  0.4em\relax IEEE, 2008, pp. 1928--1935.

\bibitem{phillips2011sipp}
M.~Phillips and M.~Likhachev, ``Sipp: Safe interval path planning for dynamic
  environments,'' in \emph{2011 IEEE International Conference on Robotics and
  Automation}.\hskip 1em plus 0.5em minus 0.4em\relax IEEE, 2011, pp.
  5628--5635.

\bibitem{chen2019crowd}
C.~Chen, Y.~Liu, S.~Kreiss, and A.~Alahi, ``Crowd-robot interaction:
  Crowd-aware robot navigation with attention-based deep reinforcement
  learning,'' in \emph{2019 International Conference on Robotics and Automation
  (ICRA)}.\hskip 1em plus 0.5em minus 0.4em\relax IEEE, 2019, pp. 6015--6022.

\bibitem{chen2017socially}
Y.~F. Chen, M.~Everett, M.~Liu, and J.~P. How, ``Socially aware motion planning
  with deep reinforcement learning,'' in \emph{2017 IEEE/RSJ International
  Conference on Intelligent Robots and Systems (IROS)}.\hskip 1em plus 0.5em
  minus 0.4em\relax IEEE, 2017, pp. 1343--1350.

\bibitem{scholler2020constant}
C.~Sch{\"o}ller, V.~Aravantinos, F.~Lay, and A.~Knoll, ``What the constant
  velocity model can teach us about pedestrian motion prediction,'' \emph{IEEE
  Robotics and Automation Letters}, 2020.

\bibitem{alahi2016social}
A.~Alahi, K.~Goel, V.~Ramanathan, A.~Robicquet, L.~Fei-Fei, and S.~Savarese,
  ``Social lstm: Human trajectory prediction in crowded spaces,'' in
  \emph{Proceedings of the IEEE conference on computer vision and pattern
  recognition}, 2016, pp. 961--971.

\bibitem{trautman2013robot}
P.~Trautman, ``Robot navigation in dense crowds: Statistical models and
  experimental studies of human robot cooperation,'' Ph.D. dissertation,
  California Institute of Technology, 2013.

\bibitem{kuderer2012feature}
M.~Kuderer, H.~Kretzschmar, C.~Sprunk, and W.~Burgard, ``Feature-based
  prediction of trajectories for socially compliant navigation.'' in
  \emph{Robotics: science and systems}, 2012.

\bibitem{chen2017decentralized}
Y.~F. Chen, M.~Liu, M.~Everett, and J.~P. How, ``Decentralized
  non-communicating multiagent collision avoidance with deep reinforcement
  learning,'' in \emph{2017 IEEE international conference on robotics and
  automation (ICRA)}.\hskip 1em plus 0.5em minus 0.4em\relax IEEE, 2017, pp.
  285--292.

\bibitem{long2018towards}
P.~Long, T.~Fanl, X.~Liao, W.~Liu, H.~Zhang, and J.~Pan, ``Towards optimally
  decentralized multi-robot collision avoidance via deep reinforcement
  learning,'' in \emph{2018 IEEE International Conference on Robotics and
  Automation (ICRA)}.\hskip 1em plus 0.5em minus 0.4em\relax IEEE, 2018, pp.
  6252--6259.

\bibitem{chen2019relational}
C.~Chen, S.~Hu, P.~Nikdel, G.~Mori, and M.~Savva, ``Relational graph learning
  for crowd navigation,'' \emph{arXiv preprint arXiv:1909.13165}, 2019.

\bibitem{socialforce}
D.~Helbing and P.~Molnar, ``Social force model for pedestrian dynamics,''
  \emph{Physical review E}, vol.~51, no.~5, p. 4282, 1995.

\bibitem{choi2012unified}
W.~Choi and S.~Savarese, ``A unified framework for multi-target tracking and
  collective activity recognition,'' in \emph{European Conference on Computer
  Vision}.\hskip 1em plus 0.5em minus 0.4em\relax Springer, 2012, pp. 215--230.

\bibitem{choi2013understanding}
------, ``Understanding collective activitiesof people from videos,''
  \emph{IEEE transactions on pattern analysis and machine intelligence},
  vol.~36, no.~6, pp. 1242--1257, 2013.

\bibitem{leal2014learning}
L.~Leal-Taix{\'e}, M.~Fenzi, A.~Kuznetsova, B.~Rosenhahn, and S.~Savarese,
  ``Learning an image-based motion context for multiple people tracking,'' in
  \emph{Proceedings of the IEEE Conference on Computer Vision and Pattern
  Recognition}, 2014, pp. 3542--3549.

\bibitem{choi2014real}
S.~Choi, E.~Kim, and S.~Oh, ``Real-time navigation in crowded dynamic
  environments using gaussian process motion control,'' in \emph{2014 IEEE
  International Conference on Robotics and Automation (ICRA)}.\hskip 1em plus
  0.5em minus 0.4em\relax IEEE, 2014, pp. 3221--3226.

\bibitem{trautman2015robot}
P.~Trautman, J.~Ma, R.~M. Murray, and A.~Krause, ``Robot navigation in dense
  human crowds: Statistical models and experimental studies of human--robot
  cooperation,'' \emph{The International Journal of Robotics Research},
  vol.~34, no.~3, pp. 335--356, 2015.

\bibitem{trautman2017sparse}
P.~Trautman, ``Sparse interacting gaussian processes: Efficiency and optimality
  theorems of autonomous crowd navigation,'' in \emph{2017 IEEE 56th Annual
  Conference on Decision and Control (CDC)}.\hskip 1em plus 0.5em minus
  0.4em\relax IEEE, 2017, pp. 327--334.

\bibitem{van2011reciprocal}
J.~Van Den~Berg, S.~J. Guy, M.~Lin, and D.~Manocha, ``Reciprocal n-body
  collision avoidance,'' in \emph{Robotics research}.\hskip 1em plus 0.5em
  minus 0.4em\relax Springer, 2011, pp. 3--19.

\bibitem{zhang2019sr}
P.~Zhang, W.~Ouyang, P.~Zhang, J.~Xue, and N.~Zheng, ``Sr-lstm: State
  refinement for lstm towards pedestrian trajectory prediction,'' in
  \emph{Proceedings of the IEEE Conference on Computer Vision and Pattern
  Recognition}, 2019, pp. 12\,085--12\,094.

\bibitem{robicquet2016learning}
A.~Robicquet, A.~Sadeghian, A.~Alahi, and S.~Savarese, ``Learning social
  etiquette: Human trajectory understanding in crowded scenes,'' in
  \emph{European conference on computer vision}.\hskip 1em plus 0.5em minus
  0.4em\relax Springer, 2016, pp. 549--565.

\bibitem{cai2019lets}
P.~Cai, Y.~Luo, A.~Saxena, D.~Hsu, and W.~S. Lee, ``Lets-drive: Driving in a
  crowd by learning from tree search,'' \emph{arXiv preprint arXiv:1905.12197},
  2019.

\bibitem{silver2017mastering}
D.~Silver, J.~Schrittwieser, K.~Simonyan, I.~Antonoglou, A.~Huang, A.~Guez,
  T.~Hubert, L.~Baker, M.~Lai, A.~Bolton \emph{et~al.}, ``Mastering the game of
  go without human knowledge,'' \emph{Nature}, vol. 550, no. 7676, pp.
  354--359, 2017.

\bibitem{sutton1992reinforcement}
R.~S. Sutton, A.~G. Barto, and R.~J. Williams, ``Reinforcement learning is
  direct adaptive optimal control,'' \emph{IEEE Control Systems Magazine},
  vol.~12, no.~2, pp. 19--22, 1992.

\bibitem{gullapalli1994acquiring}
V.~Gullapalli, J.~A. Franklin, and H.~Benbrahim, ``Acquiring robot skills via
  reinforcement learning,'' \emph{IEEE Control Systems Magazine}, vol.~14,
  no.~1, pp. 13--24, 1994.

\bibitem{tai2017virtual}
L.~Tai, G.~Paolo, and M.~Liu, ``Virtual-to-real deep reinforcement learning:
  Continuous control of mobile robots for mapless navigation,'' in \emph{2017
  IEEE/RSJ International Conference on Intelligent Robots and Systems
  (IROS)}.\hskip 1em plus 0.5em minus 0.4em\relax IEEE, 2017, pp. 31--36.

\bibitem{augugliaro2012generation}
F.~Augugliaro, A.~P. Schoellig, and R.~D'Andrea, ``Generation of collision-free
  trajectories for a quadrocopter fleet: A sequential convex programming
  approach,'' in \emph{2012 IEEE/RSJ international conference on Intelligent
  Robots and Systems}.\hskip 1em plus 0.5em minus 0.4em\relax IEEE, 2012, pp.
  1917--1922.

\bibitem{sadeghiankosaraju2018trajnet}
A.~Sadeghian, V.~Kosaraju, A.~Gupta, S.~Savarese, and A.~Alahi, ``Trajnet:
  Towards a benchmark for human trajectory prediction,'' \emph{arXiv preprint},
  2018.

\bibitem{everett2018motion}
M.~Everett, Y.~F. Chen, and J.~P. How, ``Motion planning among dynamic,
  decision-making agents with deep reinforcement learning,'' in \emph{2018
  IEEE/RSJ International Conference on Intelligent Robots and Systems
  (IROS)}.\hskip 1em plus 0.5em minus 0.4em\relax IEEE, 2018, pp. 3052--3059.

\end{thebibliography}

\end{document}